\documentclass[10pt,twocolumn,letterpaper]{article}

\usepackage{cvpr}
\usepackage{times}
\usepackage{epsfig}
\usepackage{graphicx}
\usepackage{amsmath}
\usepackage{amssymb}
\usepackage{authblk}
\usepackage{relsize}
\usepackage{tabularx, booktabs}
\usepackage{pifont}
\usepackage{enumitem}
\usepackage[table]{xcolor}

\definecolor{dartmouthgreen}{rgb}{0.05, 0.5, 0.06}
\definecolor{lightgray}{gray}{0.9}
\definecolor{lightblue}{rgb}{0.93,0.95,1.0}
\definecolor{darkgreen}{rgb}{0.0,0.6,0.0}
\definecolor{mypink1}{rgb}{0.858, 0.188, 0.478}

\newcommand{\minisection}[1]{\vspace{1mm}\noindent{\textbf{#1}}}
\newcolumntype{L}[1]{>{\raggedright\let\newline\\\arraybackslash\hspace{0pt}}m{#1}}
\newcolumntype{C}[1]{>{\centering\let\newline\\\arraybackslash\hspace{0pt}}m{#1}}
\newcolumntype{R}[1]{>{\raggedleft\let\newline\\\arraybackslash\hspace{0pt}}m{#1}}

\def\beq{\begin{equation}}
\def\eeq{\end{equation}}

\def\beqary{\begin{eqnarray}}
\def\eeqary{\end{eqnarray}}

\def\beqarz{\begin{eqnarray*}}
\def\eeqarz{\end{eqnarray*}}

\usepackage[breaklinks=true,bookmarks=false]{hyperref}

\cvprfinalcopy % *** Uncomment this line for the final submission

\def\cvprPaperID{14}

\ifcvprfinal\fi
\begin{document}

%%%%%%%%% TITLE
\title{Learn Stereo, Infer Mono:\\Siamese Networks for Self-Supervised, Monocular, Depth Estimation}

\renewcommand*{\Affilfont}{\small}
\setlength{\affilsep}{0.7em}

\author{
	Matan Goldman$^{1}$ \,\,
	Tal Hassner$^{2\dagger}$ \,\,
	Shai Avidan$^{1}$ \vspace{3pt}\\
	$^1$Tel Aviv University, Israel, $^2$The Open University of Israel, Israel}

\maketitle

\renewcommand*{\thefootnote}{$\dagger$}
\setcounter{footnote}{1}
\footnotetext{Work done while at the University of Southern California.}
\renewcommand*{\thefootnote}{\arabic{footnote}}
\setcounter{footnote}{0}
\thispagestyle{empty}

%%%%%%%%% ABSTRACT
\begin{abstract}
	The field of self-supervised monocular depth estimation has seen huge advancements in recent years. Most methods assume stereo data is available during training but usually under-utilize it and only treat it as a reference signal. We propose a novel self-supervised approach which uses both left and right images equally during training, but can still be used with a single input image at test time, for monocular depth estimation. Our Siamese network architecture consists of two, twin networks, each learns to predict a disparity map from a single image. At test time, however, only one of these networks is used in order to infer depth. We show state-of-the-art results on the standard KITTI Eigen split benchmark as well as being the highest scoring self-supervised method on the new KITTI single view benchmark. To demonstrate the ability of our method to generalize to new data sets, we further provide results on the Make3D benchmark, which was not used during training.
\end{abstract}

%%%%%%%%% BODY TEXT
\section{Introduction}

Single-view depth estimation is a fundamental problem in computer vision with numerous applications in autonomous driving, robotics, computational photography, scene understanding, and many others. Although single image depth estimation is an ill-posed problem~\cite{eigen2014depth,hassner2006example}, humans are remarkably capable of adapting to estimate depth from a single view~\cite{howard2012perceiving}. Of course, humans can use stereo vision, but when restricted to monocular vision, we can still estimate depth fairly accurately by exploiting motion parallax, familiarity with known objects and their sizes, and perspectives cues.

There is a large body of work on monocular depth estimation using classical computer vision methods~\cite{blanz1999morphable,criminisi2000single,saxena2006learning,saxena2007depth}, including several recent approaches based on convolutional neural networks (CNN)~\cite{eigen2014depth,liu2016learning}. These methods, however, are supervised and require large quantities of ground truth data. Obtaining ground truth depth data for realistic scenes, especially in unconstrained viewing settings, is a complicated task and typically involves special equipment such as light detection and ranging (LIDAR) sensors.

Several methods recently tried to overcome this limitation, by taking a self-supervised approach. These  methods exploit intrinsic geometric properties of the problem to train monocular systems~\cite{garg2016unsupervised,monodepth17}. All these cases, assume that both images are available during training, though only one training image is used as input to the network; the second image is only used as a reference. Godard et al.~\cite{monodepth17} showed that predicting both the left and the right disparity maps vastly improves accuracy. While predicting the left disparity using the left image is intuitive and straight-forward, they also estimate the right disparity using the left image. This process is prone to errors due to occlusions and information missing from the left viewpoint. By comparison, we  fully utilize both images when learning to estimate disparity from a single image.

\begin{figure*}[t]
	\centering
	\includegraphics[width=\linewidth]{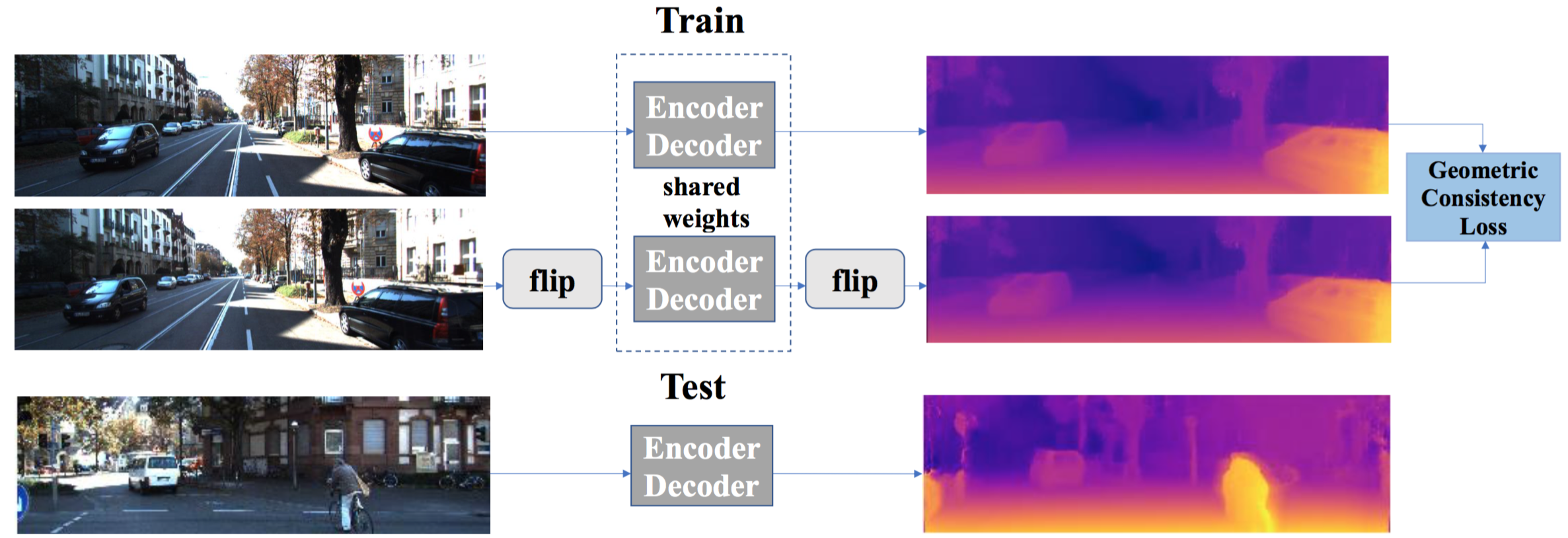}
	\vspace{0.1in}
	\caption{{\bf System overview.} Our approach uses stereo data during training, but works on single image data during test time. Both images are treated equally by mirroring the right image. We use Siamese \cite{bromley1994signature} networks with weight sharing. This reduces computational cost and allows us to run the system on single image during test time.}
	\label{fig:train_vs_test}
\end{figure*}

We propose a self-supervised approach similar to that of Godard et al.~\cite{monodepth17}. Unlike them, however, we exploit the symmetry of the disparity problem in order to obtain effective deep models. We observe that a key problem of existing methods is that they try to train a single network to predict both left and right disparity maps using a single image. This does not work well in practice since crucial information available in the right image is often occluded from the left viewpoint due to parallax (and vice versa). Instead, we propose a simple yet effective alternative approach of flipping the images around the vertical axis (vertical mirroring) and using them for training. In this way, the network only learns a left disparity map; right disparity maps are simply obtained by mirroring the right image, estimating the disparity, and then mirroring the result back to get the correct right disparity.

Specifically, we use a deep Siamese~\cite{bromley1994signature} network that learns to predict a disparity map both from the left image and the flipped right image. By using a Siamese architecture, we learn to predict each disparity map using its corresponding image. By mirroring the right image, {\em prediction of both left and right disparity maps becomes equivalent}. We can therefore train both Siamese networks using {\em shared weights}. These shared weights have the dual advantage of reducing the computational cost of training and, as evident by our results, resulting in improved networks. A high level overview of our approach is illustrated in Fig.~\ref{fig:train_vs_test}.

We evaluate our proposed system on the KITTI~\cite{Geiger2012AreSuite} and Make3D~\cite{saxena2006learning} benchmarks and show that, remarkably, in some cases our {\em self-supervised approach outperforms even supervised methods}. Importantly, despite the simplicity of our proposed approach and the improved results it offers, we are unaware of previous reports of methods which exploit the symmetry of stereo training in the same manner as we propose to do.

To summarize we provide the following contributions:
\begin{itemize}
	\item A novel approach for self-supervised learning of depth (disparity) estimation which trains on pairs of stereo images simultaneously and symmetrically.
	\item We show how a network trained on stereo images can naturally be used for monocular depth estimation at test time.
	\item We report state-of-the-art, monocular disparity estimation results which, in some cases, even outperform supervised systems.
\end{itemize}

Our code and models are available online from the following URL:~\texttt{https://github.com/mtngld/lsim}.

\section{Related work}
There is a long line of research on the problem of depth estimation. Much of this work assumed image pairs~\cite{scharstein2002taxonomy} or sequences~\cite{karsch2012depth} are available in order to infer depth. We focus on the related but different task of monocular depth estimation, where only a single image is used as input.

\minisection{Example based methods.} Example based methods use reference images with corresponding, per-pixel, ground truth depth values as priors when estimating depth for a query image. An early example is the Make3D model of Saxena et al.~\cite{saxena2006learning}, which transforms local image patches into a feature vectors and then uses a linear model trained off-line to assign depth for each query patch. These estimates were then globally adjusted using a Gaussian Markov random field (MRF). Hassner et al.~\cite{hassner2013viewing,hassner2006example,hassner2013single} suggested an on-the-fly example generation scheme which was used to produce depth estimates using a global coordinate descent method. Example based methods explicitly assume familiarity with the object classes they are being applied to. Patch based methods further have difficulties ensuring that their solutions are globally consistent.

\minisection{Scene assumption methods.} Shape-from-X methods make assumptions on the properties of the scene in order to infer depth. Some use shading in order to estimate 3D shape from a single image~\cite{basri2007photometric,horn1970shape,tuan2018extreme}. Vanishing points and other perspective cues have also been used for monocular depth estimation~\cite{criminisi2000single}. Ladicky et al.~\cite{ladicky2014pulling} suggested incorporating object semantics into the model, thus requiring additional labeled data. When objects belong to a single class, class statistics are used, as in the 3D morphable models~\cite{blanz1999morphable,tuan2017regressing}.

Other scene assumptions include the use of texture~\cite{aloimonos1988shape} and focus~\cite{nayar1994shape}. In the absence of stereo images, all these methods use visual cues inspired by human perception. Whenever these cues are absent from the scene, these approaches fail.

\minisection{Supervised, deep, monocular methods.}
Several deep learning--based methods were recently proposed for solving this problem. These methods formulated the problem using a regression function from an input image to its corresponding depth map~\cite{eigen2014depth}. Xie et al.~\cite{xie2016deep3d} used a neural network to estimate a probabilistic disparity map, followed by a selection layer. Liu et al.~\cite{liu2015deep,liu2016learning} combined the neural net approach with a conditional random field (CRF) in order to address the global nature of the problem. Roy et al.~\cite{roy2016monocular} proposed neural regression forest (NRF), a random forest method where at each tree node a shallow CNN is used. Laina et al.~\cite{laina2016deeper} trained an end-to-end fully convolutional network with residual connections and introduced
the reversed Huber loss for this task. More recently, Fu et al.~\cite{fu2018deep} suggested using ordinal regression to model this problem.

Although deep supervised methods achieve accurate results, they require large amounts of image data with corresponding ground truth depth maps. Collecting such datasets at scale is very difficult and expensive.

\begin{figure*}[t]
	\centering
	\includegraphics[clip,trim=0mm 0mm 0mm 0mm,width=0.82\linewidth]{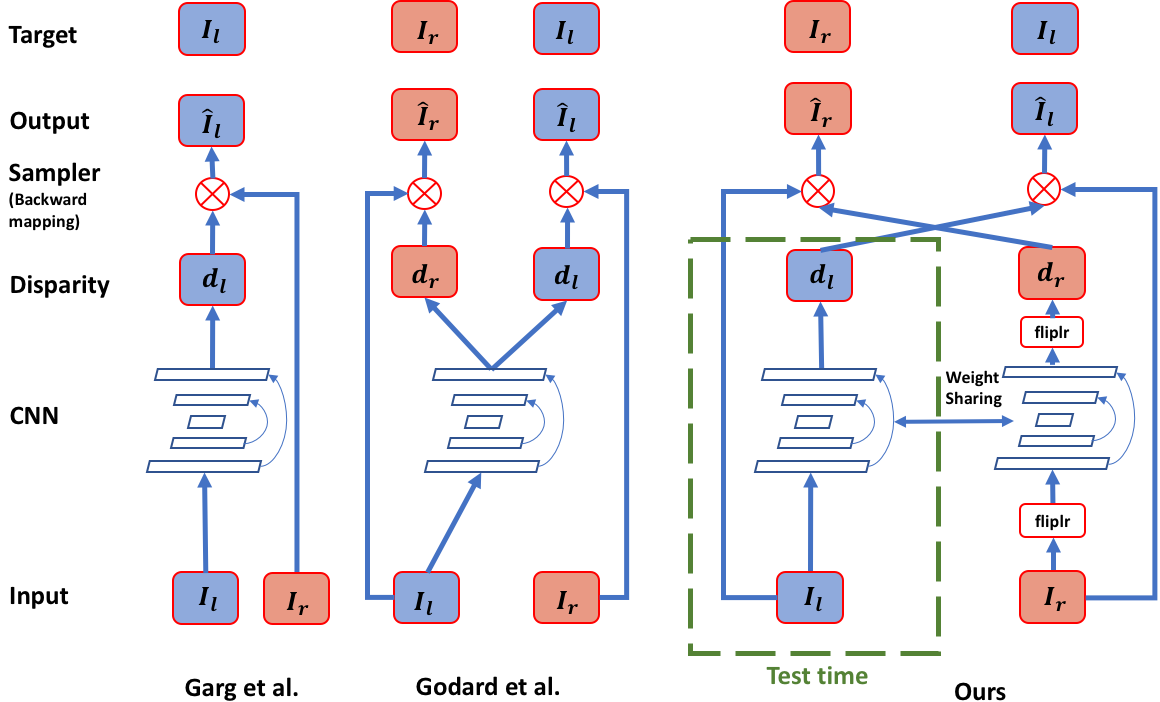}
	\caption{{\bf Comparison of system architectures.} Left: The method of Garg et al.~\cite{garg2016unsupervised} uses the right image only as a supervisory signal. Center: The method of Godard et al.~\cite{monodepth17} favors the left image over the right image. Both methods use a single image as input during training. Right: Our Siamese network trains on pairs of images, treating them both equally, by flipping the right image. Hence, our loss combines errors from two separate predictions, equally treating both views and their predictions. At test time, only the area bounded by the dashed line is used; the rest of the blocks are used only for training.}
	\label{fig:blocks}
\end{figure*}

\minisection{Self-supervised, deep, monocular methods.} Garg et al.~\cite{garg2016unsupervised} were first to suggest a self-supervised method for this problem, relying on the geometrical structure of the scene. First, they estimate a disparity image for the left image. This disparity map is then used to inverse warp the right image and measure reconstruction loss (Fig.~\ref{fig:blocks} (left)).

Our approach is related to the one recently described by Godard et al.~\cite{monodepth17}. Whereas they apply similar reasoning for data augmentation, we use a specially crafted Siamese network to better utilize the training data. Please see Sec.~\ref{mono-discussion} for a detailed discussion on the differences between their approach and ours.

Our method is further related to the one proposed by Kuznietsov et al.~\cite{kuznietsov2017semi} who also use two networks. There are some important differences between our work and theirs. First, we use two networks with weight sharing, which reduces model size and allows applying the network at test time in monocular settings. Second, they use depth information as a semi-supervisory signal. We do not use any depth information or any other labels. We report results that nearly match theirs despite the fact that our method is completely self-supervised.

Some methods suggested incorporating both depth and pose estimation~\cite{geonet,zhou2017unsupervised}. We focus solely on depth estimation and show our results to outperform the ones reported by these recent methods.
There is also a line of work \cite{kumar2018depthnet, mahjourian2018unsupervised,yang2018deep} where for using self-supervision for extracting depth from monocular video, here we do not assume sequential data is at hand.

\minisection{Siamese networks.} Siamese networks were first suggested by Bromley et al.~\cite{bromley1994signature} and have since been used for a wide range of tasks, including metric learning~\cite{chopra2005learning} and recognition~\cite{koch2015siamese}. Some recently applied Siamese networks to depth estimation~\cite{kendall2017end,luo2016efficient}. These methods were all supervised and assume stereo vision during both training and testing.

\section{Our approach}
We use pairs of RGB rectified images for training and assume the images were acquired in a controlled setup where the baseline between the cameras is known. Later on, this assumption will allow us to easily convert from disparities to depth. We believe it is reasonable to assume availability of rectified stereo pairs, even at scale, and there are several datasets containing data of this type~\cite{Cordts2016TheUnderstanding,geiger2013vision,Menze2015ObjectVehiclesb}.

We aim to learn a mapping $\hat{\mathbf{d_l}} = f(\mathbf{I_l})$, from an RGB image to a depth map and similarly $\hat{\mathbf{d}_r}=f'(\mathbf{I}_r)$. Compare to \cite{monodepth17} in which the problem during training could be formulated to $(\hat{\mathbf{d_l}},\hat{\mathbf{d_r}}) = f(\mathbf{I_l})$.

The two functions, $f()$ and $f'()$ cannot be the same: inferring a left disparity map is a different problem than inferring a right disparity map, if only because of the different relative positions of the two images and hence the different disparities that are assigned to their pixels. Clearly, we can train two separate networks, one for each function, but that would prevent weight sharing between the two networks or allow us to exploit the inherent symmetry of the problem. We propose an alternative method which utilizes both images in an equivalent manner.

\subsection{Siamese architecture with mirroring}
To make equivalent and symmetric use of available training data, we exploit the symmetry of the problem and note that by mirroring (horizontal flipping) $I_r$ we get a new image $m(I_r)$ which can be considered as being sampled from the distribution of left images, that means we can apply our $f()$ function on such image, but now, in order to return to right disparity another mirroring is required, to summarize $f'(\cdot) = m(f(m(\cdot)))$. We hence change the architecture used to train and infer depth to exploit the symmetry. These changes are presented in Figure ~\ref{fig:blocks} as a detailed block diagram of our method, compared to the designs of previous approaches. As can be seen, both Garg et al.~\cite{garg2016unsupervised} and Godard et al.~\cite{monodepth17} propose an architecture with a single input used as input during training. Garg et al. are further limited by using the right image only as a supervisory signal. We use a Siamese architecture which takes both images simultaneously as input during training, treating both views equally. Our approach therefore not only saves memory, it also shares information between the networks.

Specifically, both previous methods under-utilize the right view~\cite{garg2016unsupervised,monodepth17}: Neither feeds the right image as input to the encoder-decoder architecture. The right image is only used as reference signal to the reconstructed image $\hat{\mathbf{I}}^l(\mathbf{d}^r) = \hat{\mathbf{I}}^r$. Of course, data augmentation can be used to flip both images and present each one, separately. In doing so, however, the network cannot see regions occluded in one view but visible in the other. We discuss these limitations in detail, in Sec.~\ref{mono-discussion}.

Note that while Siamese networks require double the training time, the actual net throughput is the same as that of a single network trained separately on both images~\cite{garg2016unsupervised,monodepth17}, because two training images are viewed and processed in each step. Also note that because of weight sharing the memory consumption is also unaffected.

\subsection{Network architecture}\label{sec:architecture}
We use a network architecture based on DispNet~\cite{MIFDB16}, applying modifications similar to those described by Godard et al.~\cite{monodepth17}. We use both ResNet~\cite{he2016deep} and VGG~\cite{simonyan2014very} architecture variants. The network is composed of an encoder-decoder pair with skip connections, allowing the network to overcome data lost during down-sampling steps while still using the advantages of a deep network.

The network produces multi-scale disparity maps: $d^1_{view}, ..., d^4_{view}$ for the four scales considered by our network and $view$ representing either $l$ or $r$ for the left/right images of a stereo pair. Lower resolution disparity predictions are concatenated with previous decoder layer output and with the corresponding encoder output using the skip connections. The concatenated results are then fed into the next (higher) scale of the network~\cite{MIFDB16}. In order to warp each disparity map and image onto its counterpart, we use a bilinear sampler as in~\cite{jaderberg2015spatial} which allows for end-to-end back-propagation and learning.

\subsection{Loss function}\label{sec:loss}
We define a multi-scale loss, somewhat related to one proposed by others~\cite{monodepth17}. The single scale loss is defined by:
\begin{equation}\label{eq:lossscale}
	L_s = \alpha_{im} (L^l_{im} + L^r_{im}) + \alpha_{tv} (L^l_{tv} + L^r_{tv}) + \alpha_{lr} (L^l_{lr} + L^r_{lr}).
\end{equation}
The components of Eq.~\eqref{eq:lossscale} are defined below. Note that this loss averages prediction errors from both left and right views. This should be compared with Garg et al.~\cite{garg2016unsupervised}, who consider single view predictions, and Godard et al.~\cite{monodepth17} who average two predictions, but unlike us, their predictions are not equivalent (See also Fig.~\ref{fig:blocks}).

The total loss is then a sum over the four scales:
\begin{equation}\label{eq:loss}
	L= \sum_{s=1}^4{L_s}.
\end{equation}
We tried using only the loss defined for the most detailed (high resolution) scale but found that combining multiple scales leads to better accuracy.

An additional modification of our loss, Eq.~\eqref{eq:loss} compared with previous work~\cite{monodepth17} is that we use a {\em total variation component}, described below, instead of their disparity smoothness term. We found this change to improve disparity results. We next detail the terms included in Eq.~\eqref{eq:lossscale}.

\minisection{Image loss.} Zhao et al.~\cite{zhao2017loss} compared multiple loss functions for the task of image restoration and showed that combining $L_1$ loss with the structural similarity (SSIM) loss~\cite{wang2004image} leads to better results. It was later shown by others~\cite{monodepth17,geonet} that this loss function is very suitable for the task of depth estimation. We follow their steps and use this as our loss function. Unlike previous work~\cite{monodepth17}, however, where only an average pooling version of SSIM is applied, we use the original SSIM with a Gaussian kernel as we find it to improve the localization of the metric.

Specifically, SSIM is defined as:
\begin{equation}
	\text{SSIM}(x,y) = \frac{(2\mu_x\mu_y + c_1)(2\sigma_{xy} + c_2)}{(\mu_x^2+ \mu_y^2 + c_1)(\sigma_x^2+ \sigma_y^2 + c_2)},
\end{equation}
where $x,y$ are two equal sized windows in the two compared images. Scalars $\mu_x, \mu_y, \sigma_x$, and $\sigma_y$ are the mean and variance of $x$ and $y$ respectively, and $\sigma_{xy}$ is the covariance of $x,y$. To summarize, the image loss is therefore measured as follows:
\begin{equation}
	L^l_{im} = \frac{\alpha}{N}\sum_{i,j}\frac{1-\text{SSIM}(I^l_{ij}, \hat{I}^l_{ij})}{2} + (1 - \alpha) \| I^l_{ij} - \hat{I}^l_{ij}\|.
\end{equation}

\minisection{Left-right consistency loss.} As demonstrated by others~\cite{monodepth17}, adding a constraint on the left-right consistency of the estimated disparity images leads to improved results. Because the task we are trying to solve is self-supervised, it is reasonable to use any geometric property that can be used as feedback to the model performance. To this end, left-right consistency is introduced to the loss and defined as follows:
\begin{equation}
	L_{lr}^l = \frac{1}{N} \sum_{i,j}|d_{i,j}^l -d^r_{i,j + d_{i,j}^l}|.
\end{equation}

\minisection{Total variation loss.}
In order to promote smoothness of the estimated disparity maps we use a total variation loss that serves as a regularization term
\begin{equation}
	L_{tv}(d) = \sum_{i,j}|d_{i + 1, j} - d_{i,j}| + |d_{i, j+1} - d_{i,j}|.
\end{equation}
We have also tried weighting this loss with the gradients of the original images, as suggested by others~\cite{monodepth17}. We found, however, that this also emphasizes disparity gradients in unnecessary places in objects like windows and walls. These objects should have the same depth but have different disparities in the weighted version.

\subsection{Post-processing} \label{post-processing}
Due to occlusions, the left side of the disparity map is usually missing important data. To overcome this, we follow a post-processing method based on the one suggested by Godard et al.~\cite{monodepth17}. Given the image $I$, at test time, we also infer the depth of the horizontally mirrored image, $m(I)$. The two disparity images are later blended together using a weighting function.

\subsection{Discussion: Comparison with Godard et al.~\cite{monodepth17}} \label{mono-discussion}
It is instructional to consider the significance of the differences in the design of our approach and the related work of Godard et al.~\cite{monodepth17}.

\subsubsection{Similar loss, different components.}
As mentioned in Sec.~\ref{sec:loss}, the loss used by Godard et al. averages predictions for two views, similarly to ours. However, unlike in our approach, their predictions are not equivalent: both were produced from the left view, while the right view is used only as a supervisory signal (see also Figure ~\ref{fig:blocks}). We provide the model inputs from two views, simultaneously, treating them equally, thus the network is given more data as input and each predicted disparity map is created independently from it's corresponding image.

\minisection{Siamese Network $\neq$ Data Augmentation.} Instead of training a Siamese network, as proposed here, a single input network can be trained on the left image, with the right image used for supervision, and, separately, on the two images flipped and their roles switched~\cite{monodepth17}. This approach, however, is different than the dual-input Siamese network approach proposed here.

First, using both images allows the network to back-propagate information from one branch into the other simultaneously. This information is unavailable when training with a single view. Second, including both left and right images as input adds information which would otherwise be unavailable due to occlusions and limited field-of-view.

\begin{figure}
	\centering
	\includegraphics[width=1\linewidth]{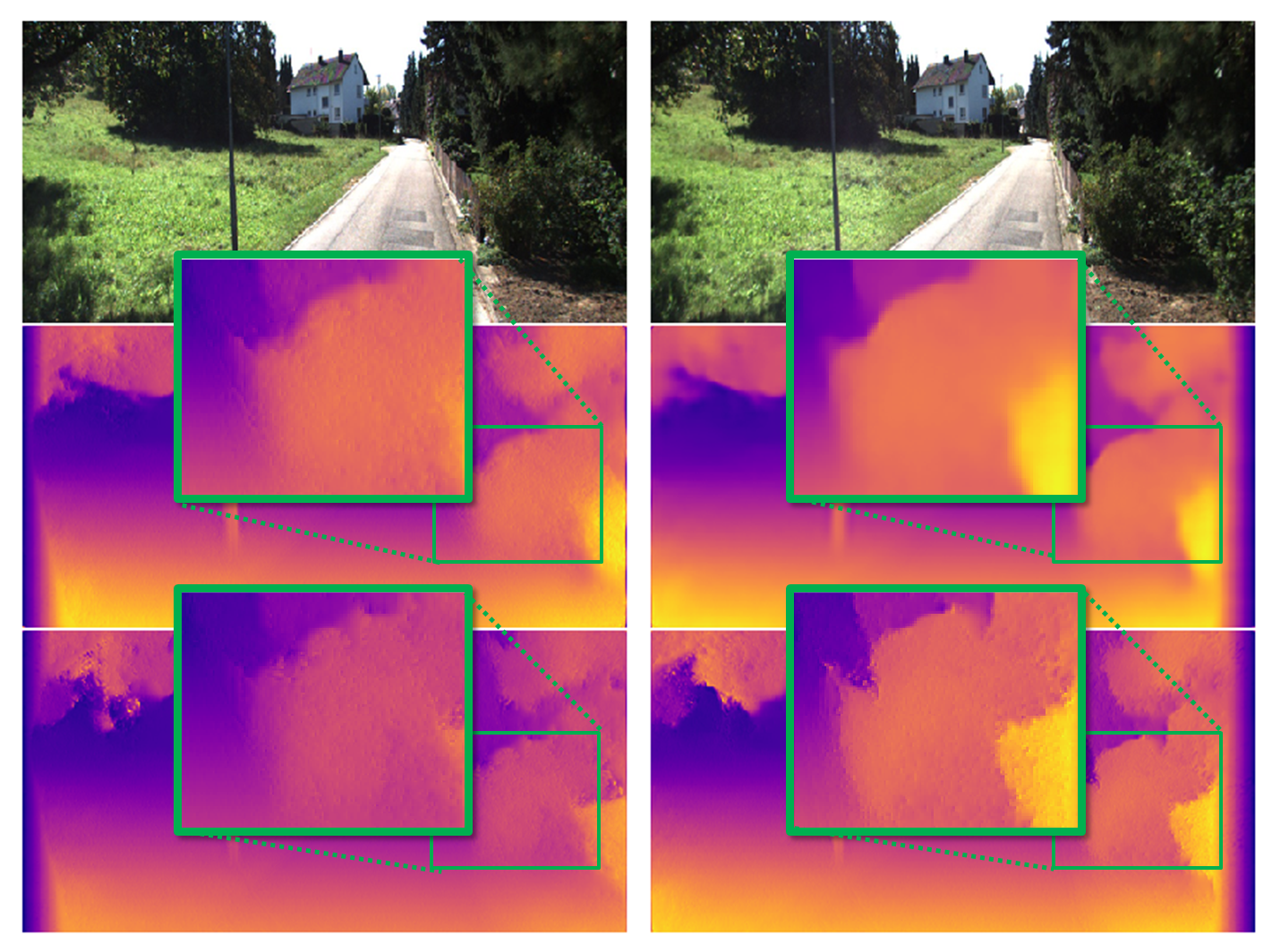}
	\caption{\textbf{Qualitative comparison of disparity maps.} Top row
		contains the input pair of images, the two rows below contains the left and right
		disparity maps predicted by Godard et al.~\cite{monodepth17} and by our method. As evident from the zoomed-in views, our results are crisper, containing more high-gradient information. This is particularly evident in depth discontinuities, such as the edge of the bushes. Also note the boundary effects, these are modeled differently for left and right disparities, hence the flipping is needed.}
	\label{fig:right_disp}
\end{figure}

Fig.~\ref{fig:right_disp}, compares the right disparity map produced by
Godard et al. to ours. Their disparity is blurry and missing important details and contours. These errors
can be intuitively explained by their uncertainty of the right image. This uncertainty creates an asymmetry between $d_l$ and $d_r$. Notice that in our prediction (bottom row) both left and
right disparities are fine grained. Put differently, the left-right consistency of our loss relies on accurate predictions of {\em both} left and right disparities. The network must therefore learn to predict the right disparity map as accurately as possible in order to minimize its loss.

\minisection{Why does flipping work? Can we just reverse the directions of the disparities?} It is possible to reverse the disparity directions, since:
$d_l = x_l - x_r$ and $d_r = x_r - x_l = - d_l$, where $x_l$ and $x_r$ are two
corresponding points in the left and right image respectively.

This approach, however, does not take into account boundary effects, as seen in Fig.~\ref{fig:right_disp}. We expect the left (right) disparity to include some boundary artifact in the left (right) side, due to missing data. Another potential limitation of this approach is that the information is distributed differently for the left and right images, $I_l \not\sim I_r$, due to the different positions of the left and right cameras. We design our network with bias towards left images, but by exploiting the symmetry and flipping right images we can assume the flipped distribution
is the same $I_l \sim m(I_r)$. This allows us to avoid bias and use the same network for both images.

\section{Results}
We tested our approach on two standard benchmark for monocular depth estimation: the KITTI Eigen split~\cite{Geiger2012AreSuite} and the KITTI single image depth prediction challenge~\cite{uhrig2017sparsity}. In addition, to show that our method generalizes well to new data, we provide results on the Make3D benchmark~\cite{saxena2006learning,saxena20083}. Importantly, Make3D has only 400 images and so training is impossibly on this set, which has appearance biases different from those of KITTI images. Our results were therefore obtained without training on Make3D images. These results are reported next.

\minisection{Implementation details.}
Similarly to previous work~\cite{monodepth17,geonet}, we first train our model on the high resolution Cityscapes dataset~\cite{Cordts2016TheUnderstanding} and later fine-tune for 30 epochs on KITTI training images~\cite{Geiger2012AreSuite}, in order to provide our network with as much training data as possible while domain-shifting to KITTI data.

For optimization we use Adam optimizer~\cite{kingma2014adam} with $\beta_1 = 0.9$, $\beta_2 = 0.999$ and $\epsilon = 10^{-8}$. We use a constant learning rate of $\lambda = 10^{-4}$. Our loss parameters of Eq.~\eqref{eq:loss} are set as: $\alpha_{im} = 1.0$, $\alpha_{lr} = 1.0$ and $\alpha_{tv} = 0.001$.

We use a batch size of eight for training. We also augment the data by applying on-the-fly color, gamma, and brightness transformations. Training uses the TensorFlow package~\cite{abadi2016tensorflow} on a Titan X GPU. The average test time for each image is 73ms. This includes processing both the image and its mirrored version.

\setlength{\tabcolsep}{3pt}

\begin{table*}[t]
	\centering
	%\scalebox{0.75}{
	\resizebox{1.0\linewidth}{!}{
		\begin{tabular}{lcccccccc}
			\toprule
			Method                                      & Dataset & Abs Rel $\downarrow$ & Sq Rel  $\downarrow$ & RMSE   $\downarrow$ & RMSE log  $\downarrow$ & $\delta < 1.25 \uparrow$ & $\delta < 1.25^2 \uparrow$ & $\delta < 1.25^3\uparrow$ \\
			\hline
			Train set mean                              & K       & 0.361                & 4.826                & 8.102               & 0.377                  & 0.638                    & 0.804                      & 0.894                     \\
			Eigen et al.~\cite{eigen2014depth} - Coarse & K       & 0.214                & 1.605                & 6.563               & 0.292                  & 0.673                    & 0.884                      & 0.957                     \\
			Eigen et al.~\cite{eigen2014depth} - Fine   & K       & 0.203                & 1.548                & 6.307               & 0.282                  & 0.702                    & 0.890                      & 0.958                     \\
			Liu et al.~\cite{liu2016learning}           & K       & 0.202                & 1.614                & 6.523               & 0.275                  & 0.678                    & 0.895                      & 0.965                     \\
			Godard et al.~\cite{monodepth17}            & CS + K  & 0.114                & \textbf{0.898}       & \textbf{4.935}      & \textbf{0.206}         & \textbf{0.861}           & \textbf{0.949}             & \textbf{0.976}            \\
			Zhou et al.~\cite{zhou2017unsupervised}     & CS + K  & 0.198                & 1.836                & 6.565               & 0.275                  & 0.718                    & 0.901                      & 0.960                     \\
			Yin et al.~\cite{geonet}                    & CS + K  & 0.153                & 1.328                & 5.737               & 0.232                  & 0.802                    & 0.934                      & 0.972                     \\
			Ours, VGG                                   & CS + K  & 0.121                & 0.9643               & 5.137               & 0.213                  & 0.846                    & 0.944                      & 0.976                     \\
			Ours, Resnet                                & CS + K  & \textbf{0.113}       & \textbf{0.898}       & 5.048               & 0.208                  & 0.853                    & 0.948                      & \textbf{0.976}            \\ \hline
			Garg et al. cap 50m~\cite{eigen2014depth}   & K       & 0.169                & 1.080                & 5.104               & 0.273                  & 0.740                    & 0.904                      & 0.962                     \\
			Yin et al.~\cite{geonet} cap 50m            & K       & 0.147                & 0.936                & 4.348               & 0.218                  & 0.810                    & 0.941                      & 0.977                     \\
			Godard et al.~\cite{monodepth17}  cap 50m   & CS + K  & 0.108                & 0.657                & \textbf{3.729}      & \textbf{0.194}         & 0.873                    & \textbf{0.954}             & \textbf{0.979}            \\
			Ours, VGG, cap 50m                          & CS + K  & 0.1155               & 0.7152               & 3.922               & 0.201                  & 0.859                    & 0.951                      & 0.979                     \\
			Ours, Resnet, cap 50m                       & CS + K  & \textbf{0.1069}      & \textbf{0.6531}      & 3.790               & 0.195                  & \textbf{0.867}           & \textbf{0.954}             & \textbf{0.979}            \\
			\bottomrule
		\end{tabular}
	}
	\vspace{0.1in}
	\caption{{\bf Results for KITTI 2015~\cite{Geiger2012AreSuite}.} Our method achieves state-of-the-art accuracy on some of the metrics and comparable results on others. Results in the top part of the table represent scenes of up to $80$ meters; the bottom part of the table provides results of up to $50$ meters. Our results follow post-processing, described in Sec.~\ref{post-processing}. Bold numbers are best.}
	\label{kitti-eigen-split}
\end{table*}

\minisection{KITTI Eigen split.}
The KITTI dataset~\cite{Geiger2012AreSuite} contains $42,382$ rectified stereo pairs from 61 scenes. Most of the images are $1,242 \times 375$ pixels in size. For easy comparison with previous work, we use the metrics and proposed train/test splits defined by others~\cite{eigen2014depth}.

KITTI Eigen split contains $697$ test images taken from 29 scenes. Additional 32 scenes are provided for training and evaluation. Ground truth depth data is created by reprojecting 3D points acquired by the Velodyne laser onto the left image. It should be noted that depth data is available only for a sparse subset of the pixels; only $5\%$ of the pixels include ground truth depth data. This ground truth data also contains measurement noise due to sensor rotation and movement of the carrying vehicle.

We use the same image crop defined by others~\cite{garg2016unsupervised}, as the same crop was used by all the baseline methods we compared with. Predictions are rescaled using bilinear interpolation in order to match the original image size. While this is the most common evaluation for the task, some concerns were recently raised regarding this methodology~\cite{godard2018digging}. We provide results for this protocol for completeness but emphasize that a more appropriate evaluation may be the KITTI single image prediction challenge~\cite{uhrig2017sparsity}, which we have also tested and for which we offer results below.

Table~\ref{kitti-eigen-split} reports results on this data set. As can be seen, our method achieves state-of-the-art accuracy in nearly all accuracy measures, with the exception of RMSE and RMSE log, where it trails the best results by a very narrow margin. Importantly, these metrics are often considered less stable.

\begin{figure*}
	\centering
	\includegraphics[width=0.90\textwidth, trim={0 0 0 3.51cm},clip]{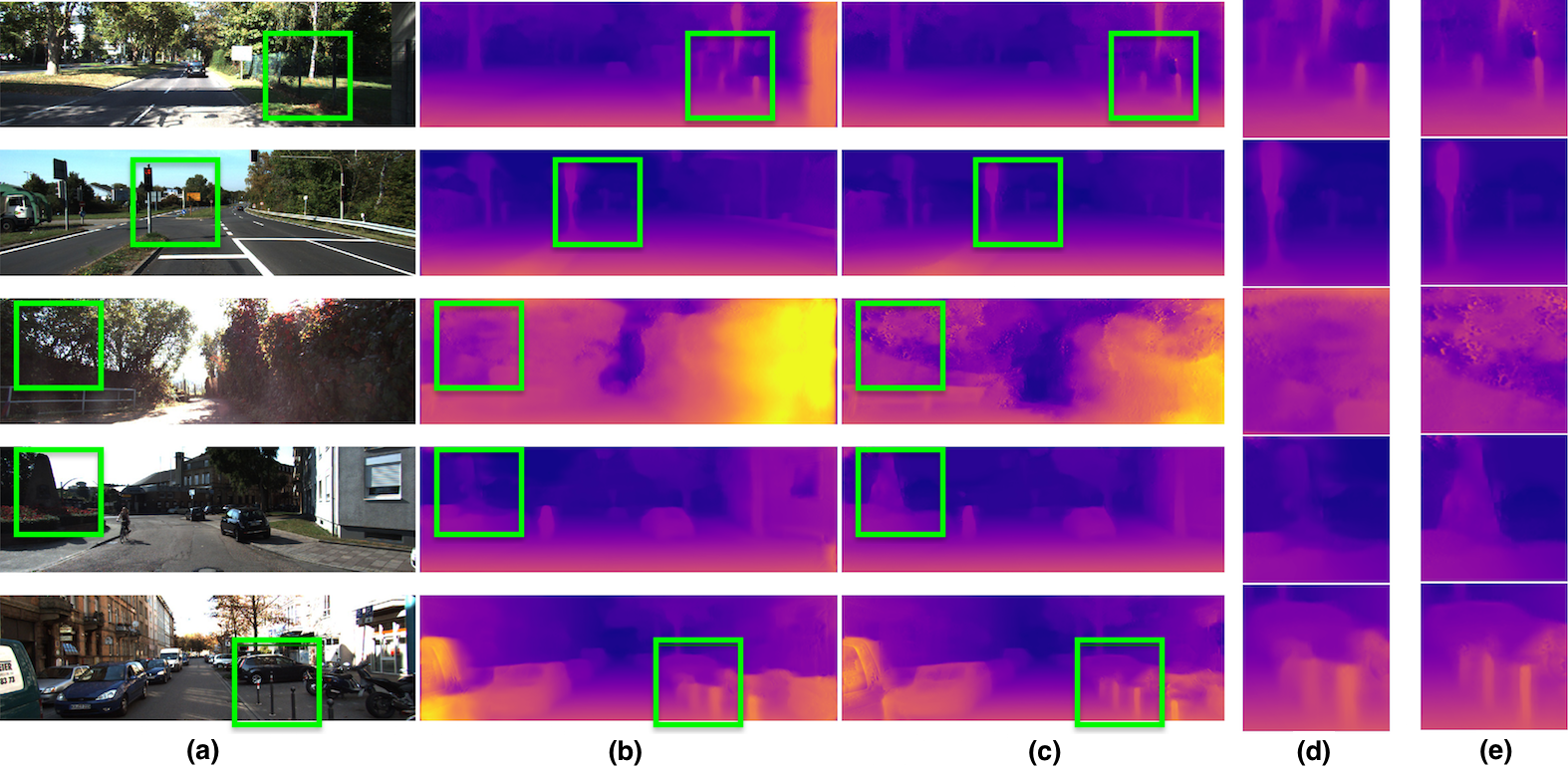}
	\caption{{\bf Qualitative comparison on KITTI data.} Comparing Godard et al.~\cite{monodepth17} (column b and zoomed-in version in column d) and our method results (column c and zoomed-in version in column e). Our method improves depth estimation for small objects and overcomes texture-less regions. For Godard et al.~\cite{monodepth17} we used a publicly available model~\cite{city2eigen_resnet}}.
\end{figure*}

\begin{figure}[t]
	\centering
	\includegraphics[width=1.0\linewidth,clip, trim={4mm 34mm 2mm 0cm}]{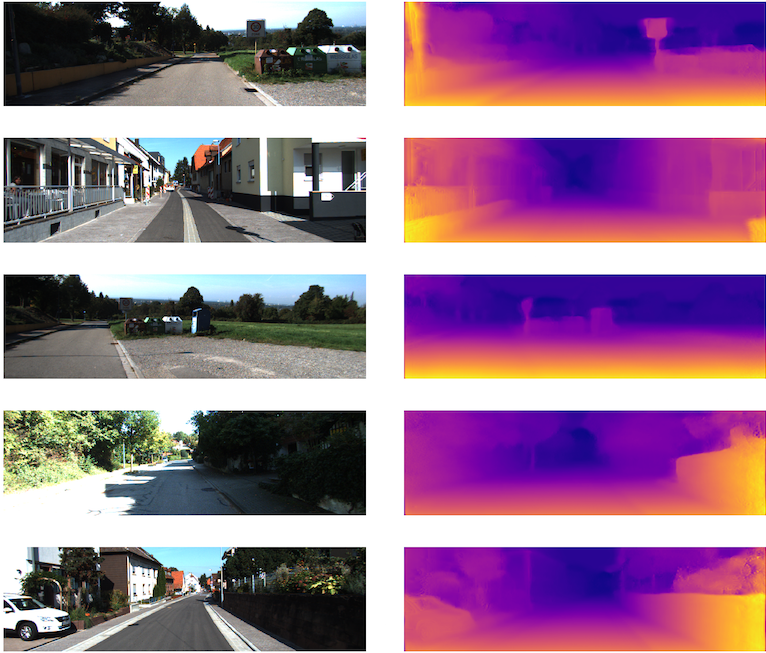}
	\caption{{\bf Qualitative disparity results on the KITTI single image depth prediction test set~\cite{uhrig2017sparsity}.} Left: RGB images. Right: Disparity maps produced by our model. Note that ground truth data is not available for these images.\vspace{-2mm}}
	\label{fig:kitti_test}
\end{figure}

\minisection{KITTI Single image depth benchmark.}
We also evaluate our method using the recently released KITTI single image depth prediction challenge~\cite{uhrig2017sparsity}. This benchmark contains 500 RGB test images that are provided for evaluation but the ground truth is only accessible to the dataset creators. We do not use the ground truth depth maps provided with the train/validation datasets. Our results are compared with existing public results in Table~\ref{kitti-single-table}, with qualitative examples of our estimates provided in Fig.~\ref{fig:kitti_test}.

As this is a fairly new challenge published by the KITTI team, there is a limited number of published results on this benchmark, all of which were obtained by supervised methods. While our method does not always achieve the best results it is the only one which is self-supervised. Still, our method achieves comparable accuracy with those supervised methods as well as outperforming the supervised baselines provided for this benchmark. In addition, our method is faster than any of these previous methods.

\begin{table}
	\centering
	\resizebox{1.0\linewidth}{!}{
		\begin{tabular}{lcccccc}
			\toprule
			Method                            & Supervision?                     & SILog          & sqRel         & absRel        & iRMSE          & Runtime         \\
			\hline
			Baseline                          & \textcolor{red}{Full}            & 18.19          & 7.32          & 14.24         & 18.50          & 0.2 s           \\
			Fu et al.~\cite{fu2018deep}       & \textcolor{red}{Full}            & \textbf{11.77} & \textbf{2.23} & \textbf{8.78} & \textbf{12.98} & 0.5s            \\
			Kong et al.~\cite{kong2018pixel}  & \textcolor{red}{Full}            & 14.74          & 3.88          & 11.74         & 15.63          & 0.2s            \\
			Li et al.~\cite{li2017monocular}  & \textcolor{red}{Full}            & 14.85          & 3.48          & 11.84         & 16.38          & 0.2s            \\
			Zhang et al.~\cite{zhang2018deep} & \textcolor{red}{Full}            & 15.47          & 4.04          & 12.52         & 15.72          & 0.2 s           \\ \hline
			Ours                              & \textcolor{dartmouthgreen}{Self} & 17.92          & 6.88          & 14.04         & 17.62          & \textbf{0.073s} \\
			\bottomrule
		\end{tabular}
	}
	\vspace{0.1in}
	\caption{{\bf Results for KITTI single image depth prediction challenge}. While the other methods are supervised our method is self-supervised yet is able to achieve comparable results. In addition, our runtime is much faster than the other listed methods. Results reported here are for the Resnet variant of our method, trained on both Cityscapes and KITTI. We note that the challenge also lists multiple unpublished methods; we report only published, non-anonymous results.}%\vspace{-5mm}
	\label{kitti-single-table}
\end{table}

\minisection{Make3D.}
In order to test the generalization of the proposed method we also evaluate it on the Make3D~\cite{saxena2006learning,saxena20083} dataset. Similarly to~\cite{monodepth17} we use a model trained only on Cityscapes data, as it is of higher resolution and contains similar scenes. We also take a central crop of the images in order to match Cityscapes aspect ratio.

The Make3D test set contains 134 pairs of single-view RGB and depth images. As common for evaluating Make3D~\cite{liu2014discrete}, we use the C1 error measures listed below, ignoring pixels where depth is larger than 70 meters:
\begin{itemize}[nosep]
	\item Squared relative error (Sq Rel): $\frac{1}{T}\sum_{i}^{T} \frac{(d_{i}^{gt} - d_{i}^{p})^2}{d_{i}^{gt}}$
	\item Absolute relative error (Abs Rel): $\frac{1}{|T|}\sum_{i}^T \frac{\|d_{i}^{gt} - d_{i}^{p}\|}{d_{i}^{gt}}$
	\item Root-mean squared error (RMSE): \tiny$\sqrt{\frac{1}{|T|}\sum_{i}^T (d_{i}^{gt} - d_{i}^{p})^2 }$\normalsize
	\item $\log_{10}$ error: $\frac{1}{|T|}\sum_{i}^T \log_{10}(d_{i}^{gt}) - \log_{10}(d_{i}^{p}) $
\end{itemize}
In all of the measures listed above, $d^{gt}_i$ and $d^p_i$ are the ground truth depth data
and the predicted depth data, respectively.

\begin{figure}
	\centering
	\includegraphics[width=1.0\linewidth,trim={0 3.1cm 0cm 0},clip]{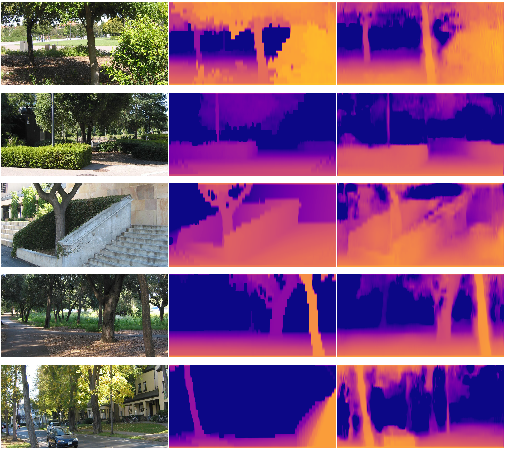}
	\caption{{\bf Qualitative results on the Make3D dataset.} (Left) Single view images used as inputs. (Center) the provided ground truth depth maps. (Right) our depth predictions as produced by a model trained on the Cityscapes dataset. As can be seen, while the quantitative results are not as good as supervised methods, the qualitative results are visually plausible.\vspace{-2mm}}
	\label{fig:make3d}
\end{figure}

We report results in Table~\ref{make3d} with some qualitative results provided in Fig.~\ref{fig:make3d}.
The strength of the proposed method is shown in its ability to perform well even when applied to a totally different domain and scene, where it outperforms other self-supervised methods and achieves comparable results to some of the supervised methods.

\begin{table}
	\centering
	\scalebox{0.83}{
		\begin{tabular}{lccccc}
			\toprule
			Method                                      & Supervision?                     & Sq Rel         & Abs Rel        & RMSE           & $\text{log}_{10}$ \\
			\hline
			Train set mean                              & \textcolor{red}{Full}            & 15.517         & 0.893          & 11.542         & 0.223             \\
			Karsch et al.~\cite{karsch2012depth}        & \textcolor{red}{Full}            & 4.894          & 0.417          & 8.172          & 0.144             \\
			Liu et al.~\cite{liu2014discrete}           & \textcolor{red}{Full}            & 6.625          & 0.462          & 9.972          & 0.161             \\
			Laina et al.~\cite{laina2016deeper}         & \textcolor{red}{Full}            & \textbf{1.665} & \textbf{0.198} & \textbf{5.461} & \textbf{0.082}    \\ \hline
			Kuznietsov et al.~\cite{kuznietsov2017semi} & \textcolor{yellow}{Semi}         & -              & 0.421          & 8.237          & 0.190             \\ \hline
			Godard et al.~\cite{monodepth17}            & \textcolor{dartmouthgreen}{Self} & 7.112          & 0.443          & 11.513         & \textbf{0.156}    \\
			Our method (Resnet)                         & \textcolor{dartmouthgreen}{Self} & \textbf{4.766} & \textbf{0.406} & \textbf{8.789} & 0.183             \\
			\bottomrule
		\end{tabular}
	}
	\vspace{0.1in}
	\caption{{\bf Comparison on the Make3D dataset}: Our method generalizes well to the unseen Make3D dataset.
		Visually, our results are plausible and consistent. Please see figure~\ref{fig:make3d} for examples. Bold numbers are best scoring for supervised and self-supervised methods respectively.}
	\label{make3d}
\end{table}

\section{Conclusions}
We propose a self-supervised method for monocular depth estimation. Our method trains on stereo image pairs but applied to to single images at test time. There is no need to provide depth information during training or any other supervisory data or labels: our system is fully self-supervised. We achieve state-of-the-art results on challenging datasets by making better use of the stereo input. Our key contribution is showing how left and right images can be symmetrically handled by mirroring the right image. Despite the simplicity of this approach, we are unaware of previous reports of similar approaches.

In addition, we provide technical contributions, including the use of a Siamese network with weight sharing for this task. As a result, we cut model size in half, using only one branch of the network at run time to process a single view input. we further define a loss function which better represents the novel design of our model.

An obvious extension of this approach is to test our method in stereo rather than monocular settings: There is nothing prohibiting our approach from being applied to stereo pairs. This ability to process monocular and stereo views is reminiscent of the human visual system which is likewise capable of generalizing from stereo to monocular settings and back. An additional direction for future work will explore the use of video and pose estimation in our suggested framework. Another technical matter that should be tackled is integrating the post-processing step into the network training
architecture to achieve a better end-to-end learning. Finally, compared to other similar systems, our approach requires relatively small networks. This small size makes it appropriate for deployment on mobile platforms.

{\small
\bibliographystyle{ieee}
%\bibliography{monostereo}

}

\end{document}